\documentclass[letterpaper, 10 pt, conference]{ieeeconf}  

\IEEEoverridecommandlockouts                              

\usepackage{amsmath,amssymb,amsfonts}
\usepackage{algorithmic}
\usepackage{graphicx}
\usepackage{textcomp}
\usepackage{xcolor}
\usepackage{multicol}
\usepackage{caption}
\usepackage{subcaption}
\usepackage[bookmarks=true]{hyperref}
\usepackage{xcolor}
\usepackage{amsmath}
\usepackage{amssymb}
\usepackage{bbm}

\PassOptionsToPackage{hyphens}{url}

\usepackage[style=ieee,dashed=false]{biblatex}

\DeclareSourcemap{
  \maps{
    \map{
      \pertype{article}
      \step[fieldset=language, null]
      \step[fieldset=url, null]
      \step[fieldset=doi, null]
      \step[fieldset=issn, null]
      \step[fieldset=isbn, null]
      \step[fieldset=note, null]
      \step[fieldset=editor, null]
      \step[fieldset=urldate, null]
      \step[fieldset=file, null]
    }
  }
}
\DeclareSourcemap{
  \maps{
    \map{
      \pertype{inproceedings}
      \step[fieldset=language, null]
      \step[fieldset=url, null]
      \step[fieldset=doi, null]
      \step[fieldset=issn, null]
      \step[fieldset=isbn, null]
      \step[fieldset=note, null]
      \step[fieldset=editor, null]
      \step[fieldset=urldate, null]
      \step[fieldset=file, null]
    }
  }
}
\DeclareSourcemap{
  \maps{
    \map{
      \pertype{incollection}
      \step[fieldset=language, null]
      \step[fieldset=url, null]
      \step[fieldset=doi, null]
      \step[fieldset=issn, null]
      \step[fieldset=isbn, null]
      \step[fieldset=note, null]
      \step[fieldset=editor, null]
      \step[fieldset=urldate, null]
      \step[fieldset=file, null]
    }
  }
}

\title{\LARGE \bf
Hierarchical RL-Guided Large-scale Navigation of a Snake Robot 
}

\author{Shuo Jiang$^{1}$, Adarsh Salagame$^{2}$, Alireza Ramezani$^{2*}$, Lawson Wong$^{1}$
\thanks{$^{1}$ The author is with the Khoury College of Computer Science, Northeastern University, Boston, MA, USA}%
\thanks{$^{2}$ The author is with the SiliconSynapse Laboratory, Department of Electrical and Computer Engineering, Northeastern University, Boston, MA, USA.}%
\thanks{{*} Corresponding author's e-mail: a.ramezani@northeastern.edu.} 
}

\begin{document}

\maketitle
\thispagestyle{empty}
\pagestyle{empty}




\begin{abstract}
Classical snake robot control leverages mimicking snake-like gaits tuned for specific environments. However, to operate adaptively in unstructured environments, gait generation must be dynamically scheduled. In this work, we present a four-layer hierarchical control scheme to enable the snake robot to navigate freely in large-scale environments. The proposed model decomposes navigation into global planning, local planning, gait generation, and gait tracking. Using reinforcement learning (RL) and a central pattern generator (CPG), our method learns to navigate in complex mazes within hours and can be directly deployed to arbitrary new environments in a zero-shot fashion. We use the high-fidelity model of Northeastern's slithering robot COBRA to test the effectiveness of the proposed hierarchical control approach.  
\end{abstract}


\section{Introduction}
Even on flat (zero elevation) ground, locomotion can be challenging in field applications due to uneven and unreliable terrain. Early research in snake robotics has focused on locomotion on flat ground and reproducing the unique locomotion of biological snakes. Typical gaits employed by snakes include lateral undulation, rectilinear motion, sidewinding, and concertina \cite{umetani_biomechanical_1974}. 

Lateral undulation is a means of movement along the longitudinal axis of the snake, relying on anisotropic friction to propel the snake forward as it moves in a sinusoidal pattern. This has been reproduced in \cite{wiriyacharoensunthorn_analysis_2002, ma_analysis_2003, ma_analysis_1999}, with most works using wheels to provide anisotropic friction for body propulsion. Snakes use rectilinear motion in tight spaces by compressing and expanding the distance between scales to produce longitudinal locomotion \cite{rincon_ver-vite_2003, ohno_design_2001}. Sidewinding gait is employed by snakes to traverse on slippery and sandy slopes. It employs a sinusoidal gait that produces lateral motion. \cite{liljeback_modular_2005, burdick_sidewinding_1994} are notable early examples of this gait implemented. Finally, the Concertina gait is used in tight spaces where the range of motion is limited by coiling and uncoiling portions of the body to generate motion in the longitudinal axis \cite{shan_design_1993}. In addition to these, non-snakelike gaits have been proposed to leverage the body's articulated nature to produce unique gaits such as the inchworm gait, slinky gait, lateral rolling gait, and tumbling locomotion \cite{yim_new_1994, rincon_ver-vite_2003}.

Research has shown that most gaits can be produced by having the snake robot's joints trace a series of sinusoids. Central pattern generators (CPG), as a dynamic oscillator, have been powerful in this regard, which can generate smoothly transitioning gaits controlled by varying the amplitude, frequency, and phases \cite{nor_cpg-based_2014, yang_hierarchical_2012, manzoor_neural_2019, wang_cpg-inspired_2017, bing_towards_2017,sihite_unsteady_2022,sihite_unilateral_2021,ramezani_generative_2021,lessieur_mechanical_2021,de_oliveira_thruster-assisted_2020,grizzle_progress_nodate,sihite_multi-modal_2023,sihite_orientation_2021,ramezani_towards_2020}.

\begin{figure}
    \centering
    \includegraphics[width=.7\linewidth]{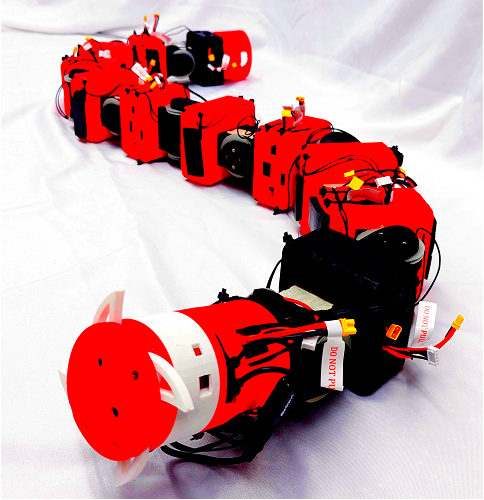}
    \caption{Illustration of Northeastern COBRA.}
    \label{fig:cover-image}
\end{figure}

Using a fixed gait can only be applied to specific simple terrains. When the snake robot is deployed on unknown complex terrains, real-time gait adjustment based on environmental perception becomes the key to navigation. Model Predictive Control (MPC) has also been used to generate dynamic gaits that can avoid obstacles and minimize energy consumption \cite{hannigan_automatic_2020, nonhoff_economic_2019, marafioti_study_2014, muller_economic_2020, fukushima_model_2021}. However, an accurate dynamic model of the snake robot being in contact with the terrain is hard to obtain due to the characteristics of rich-contact. As a model-free control method, Reinforcement learning (RL) has received extensive attention in recent years and has been successfully applied to snake robots \cite{liu2022reinforcement, shi2020deep, jia2021coach, bing2020energy}. RL has the disadvantage that requires large amount of environment interaction, and directly applying RL methods to robots with a continuous action space with high degrees of freedom is inefficient. Therefore, the method of hierarchical RL has been proposed, and the combination of CPG and RL is one of them \cite{bellegarda2022cpg, liu2022reinforcement, liu2021learning}. RL is only responsible for the parameter adjustment of CPG, and CPG generates the gait that controls the robot joints. However, such hierarchical control scheme was only verified by simple tasks such as straight path tracking.

In this work, we aim to combat the challenge of locomotion control of high-DoF slithering robots by RL-CPG control scheme to enable the robot freely navigate in large-scale complex environments. We improved the previous RL-CPG method in the following aspects: first, adding a global planning module enables the robot to be deployed in new environments without retraining. Second, the robot uses ego-centric perception in training, which can be collected directly from on-body sensors without using external measuring tools, such as complex motion tracking systems. The latter is less applicable in outdoor environments. In addition, the global planning module decomposes complex navigation problems into reusable simple local navigation problems, significantly reducing the problem's complexity and accelerating the training of RL.

\section{COBRA's Hardware Overview}
\begin{figure*}
    \centering
    \includegraphics[width=0.6\linewidth]{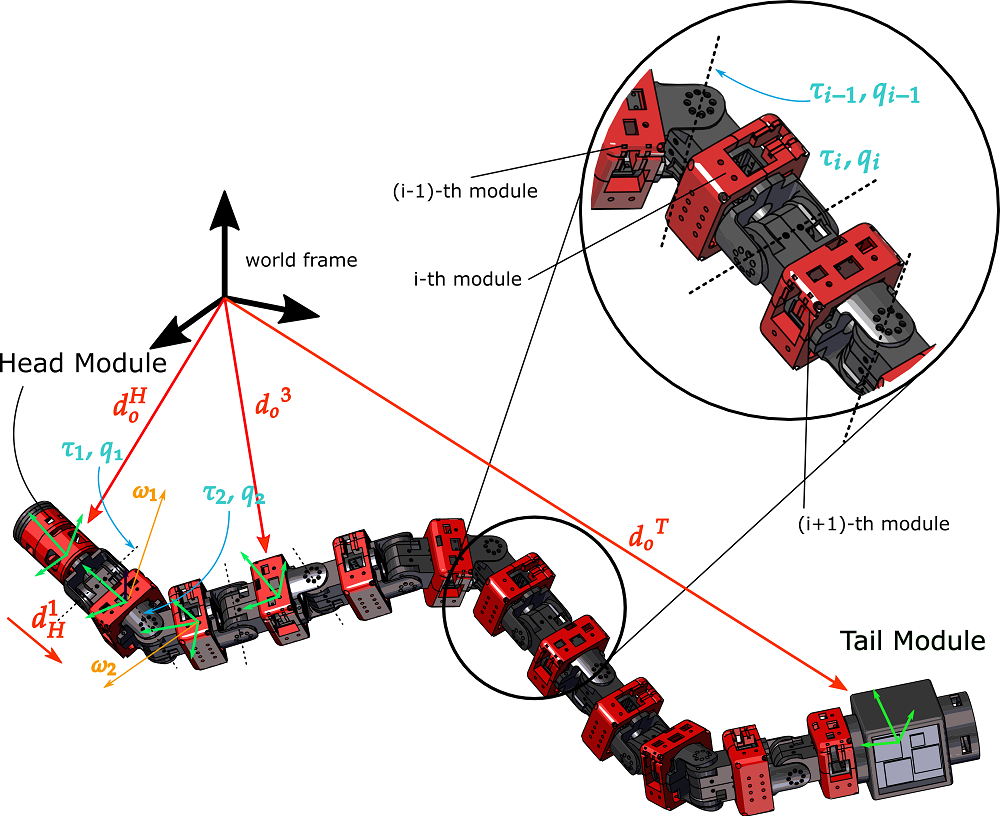}
    \caption{Illustrates the coordinate frames and parameters used for modeling COBRA.}
    \label{fig:cobra-coordinates}
\end{figure*}

COBRA, shown in Fig.~\ref{fig:cobra-coordinates}, is designed for space explorations \cite{wugiasoihf}. 
It consists of eleven actuated joints. The front of the robot consists of a head module containing the onboard computing of the system, a radio antenna for communicating with a lunar orbiter, and an inertial measurement unit (IMU) for navigation. At the tail end, there is an interchangeable payload module containing a neutron spectrometer to detect water ice for our mission. The rest of the system consists of identical 1-DoF joint modules (Fig.~\ref{fig:cobra-coordinates}) containing a joint actuator and a battery.

\subsection{Head-tail Locking Mechanism}
In addition to the eleven identical modules, COBRA features a distinct module at the snake’s head, aptly referred to as the ``head module,” and, similarly, a ``tail module” at the snake’s tail end. The head module is shown in Fig.~\ref{fig:cover-image}. The primary purpose of these unique modules is to connect to form a loop before the onset of a tumbling mode, a mode used for rapid mobility on lunar craters. The head module acts as the male connector and utilizes a latching mechanism to sit concentrically inside the female tail module. 

The latching mechanism consists of a Dynamixel XC330 actuator, which sits within the head module and drives a central gear. This gear interfaces with the partially geared sections of four fin-shaped latching ``fins.” The curved outer face of each latching fin has an arc length equal to 1/4 of the circumference of the head module’s circular cross-section. When the mechanism is retracted, these four fins form a thin cylinder that coincides with the head module's cylindrical face. A dome-shaped cap lies on the end of the head module so that the fins sit between it and the main body of the head module. Clevis pins are used to position the fins in this configuration. COBRA’s tail module features a female cavity for the fins. When transitioning to tumbling mode, the head module is positioned concentrically inside the tail module using the joint’s actuators, and the fins unfold into the cavity to lock the head module in place. For the head and tail modules to unlatch, the central gear rotates in the opposite direction, and the fins retract, allowing the system to return to sidewinding mode.

The choice for an active latching mechanism design stemmed from the design requirements and restrictions. Magnets were initially discussed as a passive latching option; however, they would not be effective in conjunction with the ferromagnetic regolith. Further, due to the need to stay in a latched configuration even when a large amount of force is applied to the system during tumbling, a passive system was not chosen, for there would be the risk of unlatching during tumbling. 
\section{Modeling}
This section aims to establish a numerical framework to study the behavior of COBRA. 

\subsection{Kinematics}
Consider the configuration variable vector $q=[\dotsc, q_i, \dotsc, p_H^\top, q_H^\top]^\top$ which embodies the body angles $q_i$, head module position $p_H$, and orientations $q_H$. We use the Euler convention to find the rotation matrix $R_H^0$  
\begin{equation}
R_{H}^{0} (q_H) =R_{z, q_z} R_{y, q_y} R_{x, q_x} 
\label{eq:head-base-transformation}
\end{equation}
\noindent $R_H^0$ represents the points in the head frame with respect to the world frame. We consider the rotation matrices $R_{i}^{H}$ 
\begin{equation}
R_{i}^{H} =R_{H}^{0}R_{i}^{H},~ i=1,\dots,N
\end{equation}
\noindent $R_i^H$ gives the expression of any points at each module with respect to the head frame. Using $R_{i}^{H}$ and $R_{H}^{0}$, we obtain the forward kinematics equation and find the center of mass (CoM) position $p_{cm,i}$ of each module with respect to the world frame
\begin{equation}
p_{cm,i}=R_{i}^{0} p_{cm,i}^{i}+d_{i}^{0}
\end{equation}
\noindent In this equation, $d_{i}^{0}$ denotes the world frame position of the body coordinates attached to i-th module.
Angular velocity of the head module $\omega_H(t)$ and its relationship with the time derivative of the configuration variable $\dot q$ is given by
\begin{equation}
\hat\omega_H(t) = \dot{R}_H^0(t)R_H^0(t),~ \omega_H=\beta_H(q)\dot q
\end{equation}
\noindent $\beta_H(q)$ is the head-module Jacobian matrix. The angular velocity of i-th module and its relation to $\dot q$ are given by
\begin{equation}
\hat\omega_i(t) = \dot{R}_i^0(t)R_i^0(t),~ \omega_i=\beta_i(q)\dot q
\end{equation}
\noindent The world-frame velocity of i-th module CoM $v_{cm,i}$ can be obtained by
\begin{equation}
v_{cm,i}^{0}=\hat\omega_i(t)R_{i}^{0} p_{cm,i}^{i}+\dot{d}_i^0=\sum_{i=1}^N \frac{\partial d_i^0}{\partial q_i} \dot{q}_i
\end{equation}

\subsection{Euler-Lagrange Formalism and Contact Model}

The dynamical equations of motion of COBRA are given by
\begin{equation}
    \begin{aligned}
        \left[\begin{array}{cc}
D_H & D_{H a} \\
D_{a H} & D_a
\end{array}\right]\left[\begin{array}{l}
\ddot{q}_H \\
\ddot{q}_a
\end{array}\right]&+\left[\begin{array}{l}
H_H \\
H_a
\end{array}\right]=\left[\begin{array}{l}
0 \\
B_a
\end{array}\right]u+\\&
+
\left[\begin{array}{cc}
J_{H}^\top \\ J_{a}^\top
\end{array}\right]F_{GRF}
    \end{aligned}
\label{eq:full-dynamics}
\end{equation}
\noindent where $D_i$, $H_i$, $B_i$, and $J_i$ are partitioned model parameters corresponding to the head 'H' and actuated 'a' joints. $F_{GRF}$ denotes the ground reaction forces. $u$ embodies the joint actuation torques. The ground model used in our simulations is given by 
\begin{equation}
\begin{aligned}
    F_{GRF} &= \begin{cases} \, 0 ~~  \mbox{if } p_{C,z} > 0  \\
     [F_{GRF,x},\, F_{GRF,y},\, F_{GRF,z}]^\top ~~ \mbox{else} \end{cases} \\
    F_{GRF,z} &= -k_1 p_{C,z} - k_{2} \dot p_{C,z} \\
    F_{GRF,i} &= - s_{i} F_{GRF,z} \, \mathrm{sgn}(\dot p_{C,i}) - \mu_v \dot p_{C,i} ~~  \mbox{if} ~~i=x, y\\
    s_{i} &= \Big(\mu_c - (\mu_c - \mu_s) \mathrm{exp} \left(-|\dot p_{C,i}|^2/v_s^2  \right) \Big)
\end{aligned}
\end{equation}
\noindent where $p_{C,i},~~i=x,y,z$ are the $x-y-z$ positions of the contact point; $F_{GRF,i},~~i=x,y,z$ are the $x-y-z$ components of the ground reaction force assuming a point contact takes place between the robot and the ground substrate; $k_{1}$ and $k_{2}$ are the spring and damping coefficients of the compliant surface model; $\mu_c$, $\mu_s$, and $\mu_v$ are the Coulomb, static, and viscous friction coefficients; and, $v_s > 0$ is the Stribeck velocity. 

While active posture control of COBRA in simulation and hardware is currently possible, we keep the scope of this work limited to simulations. The next steps involve projecting our control framework to the physical hardware, which will be reported in the sequel papers. Next, we outline our reinforcement-learning-based locomotion control of COBRA.

\section{Control}
In this section, first, we present background on the Markov decision process (MDP) and central pattern generators (CPGs) before we outline the hierarchical control approach.

\subsection{Markov Decision Process (MDP)}
A MDP is a 4-tuple $M=\left \langle S,A,P,R \right \rangle$ where $S$ is the set of states, $A$ is the set of actions, $P\left ( s_{t+1}| s_{t}, a_{t}\right )$ is the transition probability that action $a$ in state $s$ at time $t$ that will lead to state $s$ at time $t+1$, $R\left ( a_{t},s_{t} \right )$ is the distribution of reward when taking action $a_{t}$ in state $s_{t}$. A policy $\pi\left ( a_{t} |s_{t} \right )$ is defined as the probability distribution of choosing action $a_{t}$ given state $s_{t}$. The learning goal is to find a policy $\pi ^{*}$ that maximizes the accumulated reward in given horizon $T$, $\pi ^{*}=\underset{\pi }{\textup{argmax}}\underset{a_{t},s_{t}\sim \pi}{\mathbb{E }}\left [ \sum _{t=0}^{T-1}\gamma^{t}\cdot R\left ( a_{t},s_{t} \right ) \right ]$, where $\gamma$ is discount factor. Reinforcement Learning (RL) algorithms are common choices to solve MDP problems.

\subsection{Central Pattern Generators (CPGs)}
CPG is a neural circuit in the vertebrate spinal cord that generates coordinated rhythmic output signals to control robot locomotion. CPG-based control methods have been successfully applied to many kinds of robots, such as multi-legged robots \cite{bellegarda2022cpg, kent2022improved, bellegarda2022visual} or snake robots \cite{liu2020learning, bing2017cpg, liu2021learning}. Usually, to improve the terrain adaptability of CPG, optimization algorithms are often applied to adjust CPG parameters in real-time. As multiple CPG structures have been proposed, we adopted the structure in \cite{bing2017cpg}. The dynamics of CPG are shown in Equation \ref{eq_1}.
\begin{equation}
    \begin{split}
         \dot{\varphi}=\omega +\textbf{A}\cdot \varphi +\textbf{B}\cdot \theta \\
         \ddot{r}=a\cdot \left [ \frac{a}{4} \left ( R-r \right )-\dot{r}\right ]\\
         x=r\cdot sin\left ( \varphi  \right )+\delta 
    \end{split}
    \label{eq_1}
\end{equation}
\begin{equation}
    \textbf{A}=\begin{bmatrix}
 -\mu _{1}& \mu _{1} &  &  &  \\
 \mu _{2}& -2\mu _{2} & \mu _{2} &  &  \\
 &  & \ddots  &  &  \\
 &  & \mu _{n-1} & -2\mu _{n-1} & \mu _{n-1} \\
 &  &  & \mu _{n} & -\mu _{n} \\
\end{bmatrix}
\end{equation}
\begin{equation}
    \textbf{B}=\begin{bmatrix}
1 &  &  &  &  \\
-1 & 1 &  &  &  \\
 & -1 & \ddots  &  &  \\
 &  &\ddots   &1  &  \\
 &  &  &-1  &  1\\
 &  &  &  &  -1\\
\end{bmatrix}
\end{equation}
$\varphi \in \mathbb{R} ^{n}$ and $r \in \mathbb{R} ^{n}$ are internal states of CPG, $n$ is the number of output channels, usually equals the number of robot joints. $a$ and $\mu_{i}$ are hyperparameters that control the convergence rate. $R\in \mathbb{R} ^{n}$, $\omega\in \mathbb{R} ^{n}$, $\theta\in \mathbb{R} ^{n-1}$, $\delta\in \mathbb{R} ^{n}$ are inputs that control the desired amplitude, frequency, phase shift and offset. $x \in \mathbb{R} ^{n}$ is the output sinusoidal waves of $n$ channels.

\subsection{Large-scale Navigation Scenario}
\begin{figure}[t]
\begin{subfigure}{.5\linewidth}
    \centering
    \includegraphics[height=.13\textheight]{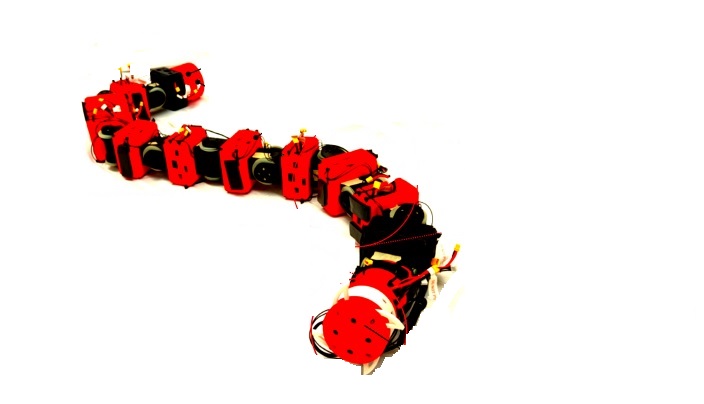}
    \caption{}
    \label{fig_4}
\end{subfigure}
\begin{subfigure}{.4\linewidth}
    \centering
    \includegraphics[height=.14\textheight]{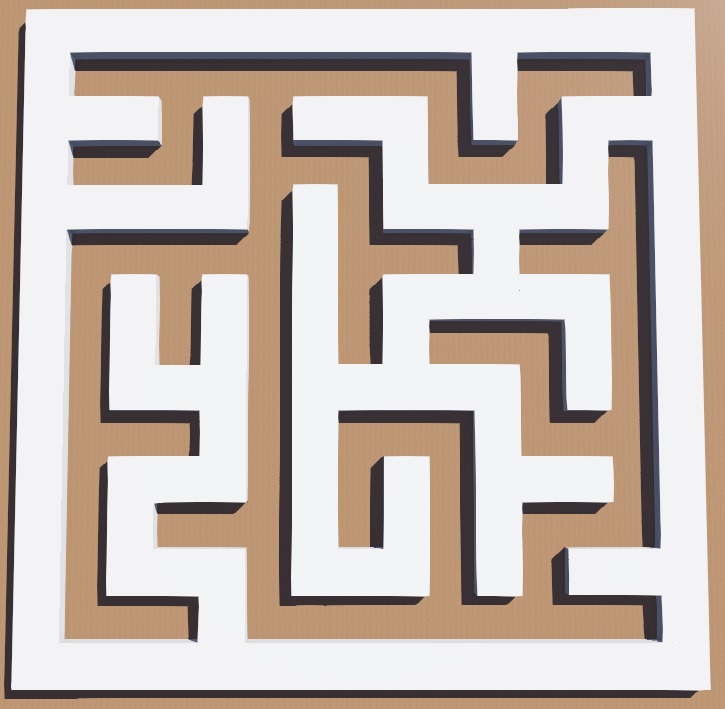}
    \caption{}
    \label{fig_3}
\end{subfigure}
\caption{Large-scale navigation scenario considered in this paper. For a goal point in a maze-shaped world like this figure, COBRA performs path planning and tracking, including slithering in a straight path and making turns at junctions.}
\end{figure}

In this work, we address the problem of COBRA navigation in large-scale scenarios. In the complex scene shown in Figure \ref{fig_3}, COBRA can move from any initial position to any goal position through locomotion. The controller is designed to directly adapt to new scenes in a zero-shot manner, given the scene map, without retraining. The COBRA robot in Fig.~\ref{fig_4} has sensors that include joint encoders and an IMU in the head to measure linear accelerations. At the same time, we provide the relative pose between the robot and the target.
\subsection{Four-layer Hierarchical Control Scheme}
\begin{figure}[t]
\centering
\includegraphics[width=.9\linewidth]{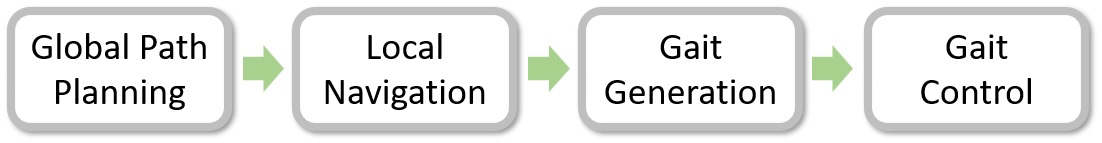}
\caption{Four-layer hierarchical control scheme.}
\label{fig_1}
\end{figure}

We designed a four-layer hierarchical control scheme to enable the snake robot with dynamics given by Eq.~\ref{eq:full-dynamics} to complete the above task. The controller structure is shown in Fig.~\ref{fig_1}. At the topmost level, global path planning uses tree search to find a path from the robot to the goal position and decomposes it into a series of waypoints. By adjusting the CPG parameters, local navigation controls the robot's movement from one waypoint to another. Gait generation converts the CPG parameters into CPG output signals to provide the target positions of the robot joints. The lowest level of gait control uses motor feedback control to track the target gait signal. 

\subsubsection{Global Path Planning}
\begin{figure}[t]
\begin{subfigure}{.45\linewidth}
    \centering
    \includegraphics[height=.10\textheight]{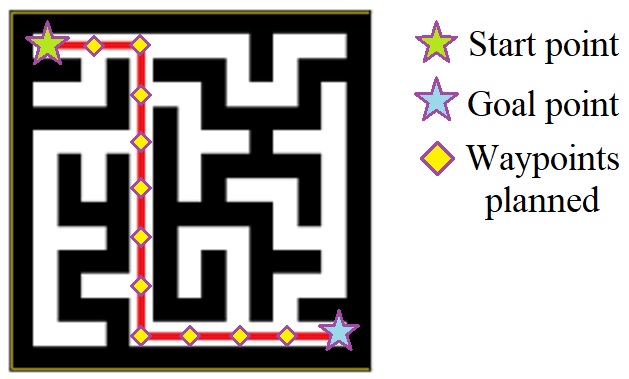}
    \caption{}
    \label{fig_2}
\end{subfigure}
\begin{subfigure}{.5\linewidth}
    \centering
    \includegraphics[height=.10\textheight]{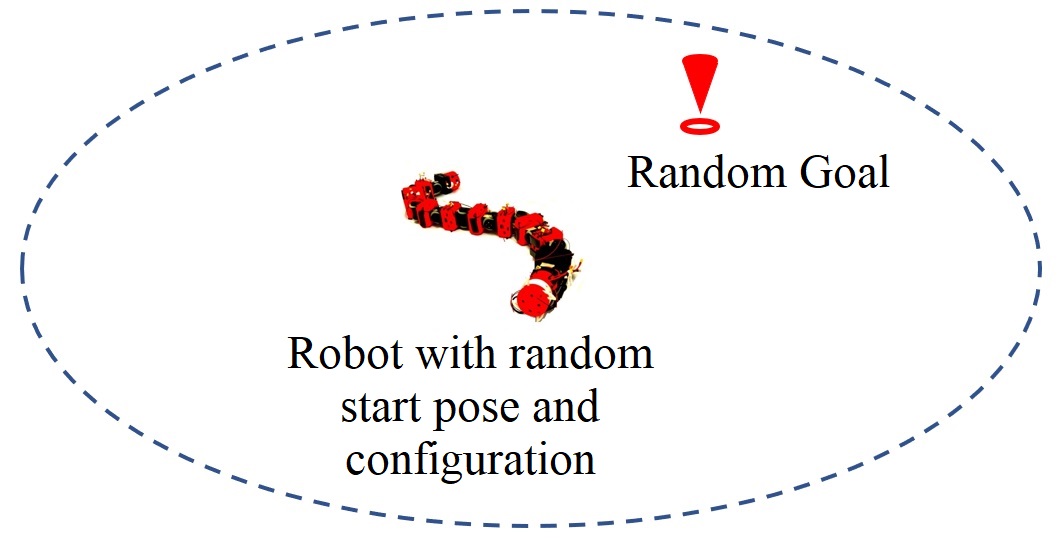}
    \caption{}
    \label{fig_8}
\end{subfigure}
\caption{For given start and goal points, the RL-guided locomotion control proposed in this work steers COBRA's eleven joints to reach the destination point.}
\end{figure}

Global path planning aims to find the shortest path in the domain, routing the robot to the destination. Given an occupancy grid of the domain, we use a tree search algorithm (A* algorithm particularly) to find the shortest path. We set several waypoints along the planned path with any two consecutive waypoints having a distance within a specific range (Fig.~\ref{fig_2}). Then the problem turns out to be how to navigate the robot in a limited area to any goal (Fig.~\ref{fig_8}). This setting dramatically reduces the complexity of the whole task compared with direct training to navigate the entire domain. The latter can be slow for training and less transferable. The local navigation task will be delivered to the lower-level RL controller.

\subsubsection{Local Navigation}
Due to the nature of snake-like locomotion (i.e., contact-rich problems), the description of its dynamic equation is extremely complicated; that is, the acyclic contact occurrence in Eq.~\ref{eq:full-dynamics} renders the control problem confounding. Most previous works adopted a simplified dynamics model and only assumed flat or regular ground. To improve the generalization of our method, we employed RL to solve the problem of local navigation proposed in the previous subsection 
 and shown in Fig.~\ref{fig_8}.

\textbf{State space:} the joint positions $\mathbb{R}^{n}$, IMU readings $\mathbb{R}^{3}$, spatial displacement between robot frame and waypoint frame $\mathbb{R}^{3}$, relative rotation between robot frame and waypoint frame are parameterized by axis-angle system $\mathbb{R}^{4}$, i.e., 21 DOFs in total. Notice the difference between our kinematics parameterization and previous works. We only use ego-centric observations from the robot, where a complex motion capture system is not required as in \cite{transeth20083, liljeback2011experimental, hasanzadeh2010ground}.

\textbf{Action space:} the action controls the parameter of two CPG components (see Gait Generation) of dimension $\mathbb{R}^{7}$. The action ranges can be summarized as (1) amplitude $R_{1}, R_{2}\in \left [ 0,1.5 \right ]$, (2) frequency $\omega _{1}=\omega _{2}\in \left [ -0.1,0.1 \right ]$, (3) phase shift $\theta  _{1},\theta _{2}\in \left [ -\pi,\pi  \right ]$, and (4) offset $\delta  _{1},\delta _{2}\in \left [ -0.1,0.1  \right ]$. 

\textbf{Reward:} We encourage the robot to reach the waypoint as soon as possible. The reward consists of the following terms:
\begin{equation}
    \begin{split}
        r_{1} &= \frac{1}{0.1+d_{t}}\\
        r_{2} &= d_{t-1}-d_{t}\\
        r_{3} &= \left\|a_{t}-a_{t-1} \right\|_{2}
    \end{split}
\end{equation}
\noindent where $d_{t}$ is the displacement between the robot frame and the waypoint frame. $r_{1}$ rewards nearer relative position and $r_{2}$ encourages approaching velocities. $r_{1}$ and $r_{2}$ work complementarily while $r_{1}\rightarrow 0$ when robot is far away from the goal and $r_{2}\rightarrow 0$ when robot is near the goal. $r_{3}$ ensures a smooth gait transition that penalizes the change of CPG parameters in consecutive planning steps.
\begin{figure}[t]
\centering
\includegraphics[width=.6\linewidth]{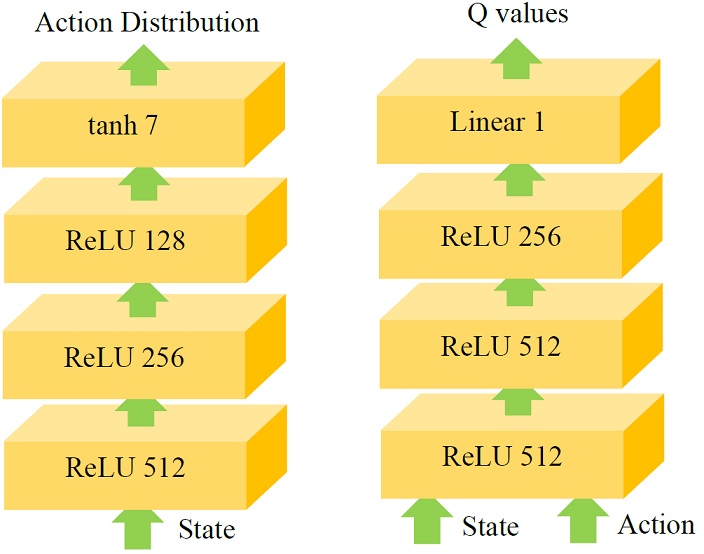}
\caption{Neural network structure of DDPG.}
\label{fig_10}
\end{figure}

\textbf{Training Details:} 
Since CPG produces a rhythmic gait, we need to reserve a certain amount of time for it to perform an effective gait, so RL should not change the CPG parameters too frequently. In this project, we let RL algorithm learn at a frequency of 0.5Hz, while the CPGs generate output signals at 50Hz to control the motors. Compared with the previous works where the learning frequency of RL is equal to the output frequency of CPG, or directly performing RL in the joint space, this design enables the RL algorithm to run more efficiently by reducing the rolling out length. Each episode lasts 160 seconds and a new goal will respawn in a random location in 8m$\times$8m squared area. The total number of episodes is 40k that can be trained in a few hours. DDPG algorithm \cite{lillicrap2015continuous} is adopted as the backbone of RL, and the network structure is shown in Fig.~\ref{fig_10}.
\begin{figure}[t]
\centering
\includegraphics[width=.95\linewidth]{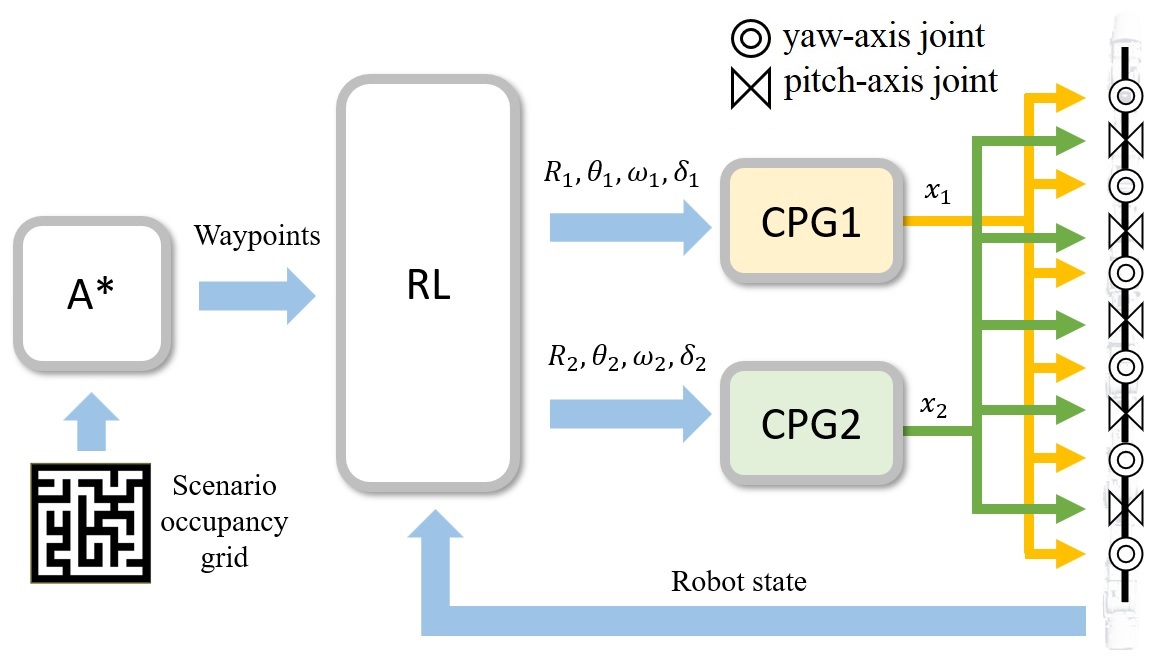}
\caption{RL block takes in the waypoints planned by the high-level global path planning component and the robot state, and it predicts the parameter of the two CPGs to guide the movement of the yaw-axis and pitch-axis joints.}
\label{fig_7}
\end{figure}
\subsubsection{Gait Generation}
We use 2 CPGs to control both pitch and yaw-axis joints (Fig.~\ref{fig_7}). Each CPG has 4 parameters to control its output sinusoidal waves, amplitude $R\in \mathbb{R} ^{k}$, frequency $\omega\in \mathbb{R} ^{k}$, phase shift $\theta\in \mathbb{R} ^{k-1}$ and offset $\delta\in \mathbb{R} ^{k}$. $k$ is the number of channels a CPG governs. To generate a rhythmic movement pattern, we set each parameter of the four mentioned with one single value. Also, with prior knowledge, valid gait emerges only when the two CPGs have the same frequency.



\subsubsection{Gait Control}
For low-level motor control, we used PID controller for each joint to track the joint positions outputted by CPGs. The PID control generates the input torque $u$ that steers the states in the model given by Eq.~\ref{eq:full-dynamics}.

\section{Simulation Results}
\label{sec:result}
\begin{figure}[t]
    \centering
    \includegraphics[width=1\linewidth]{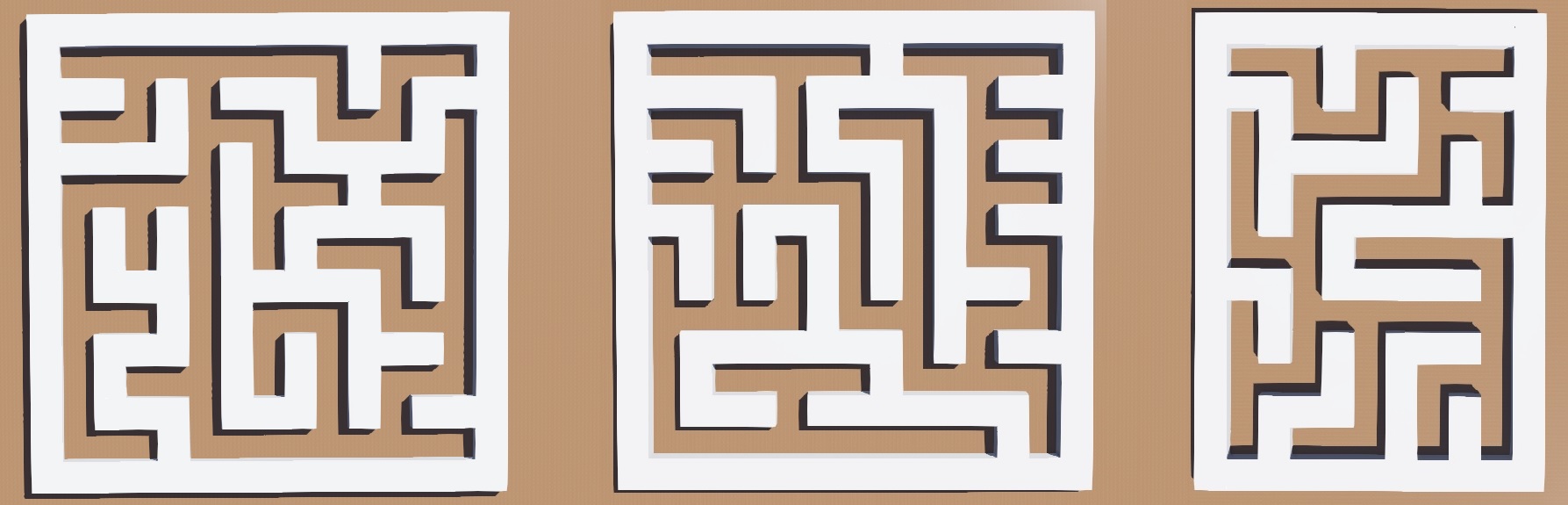}
    \caption{Randomly generated maze layouts and sizes. The learned control scheme can be directly deployed without retraining.}
    \label{fig_23}
\end{figure}

\begin{figure}[t]
    \centering
    \includegraphics[width=1\linewidth]{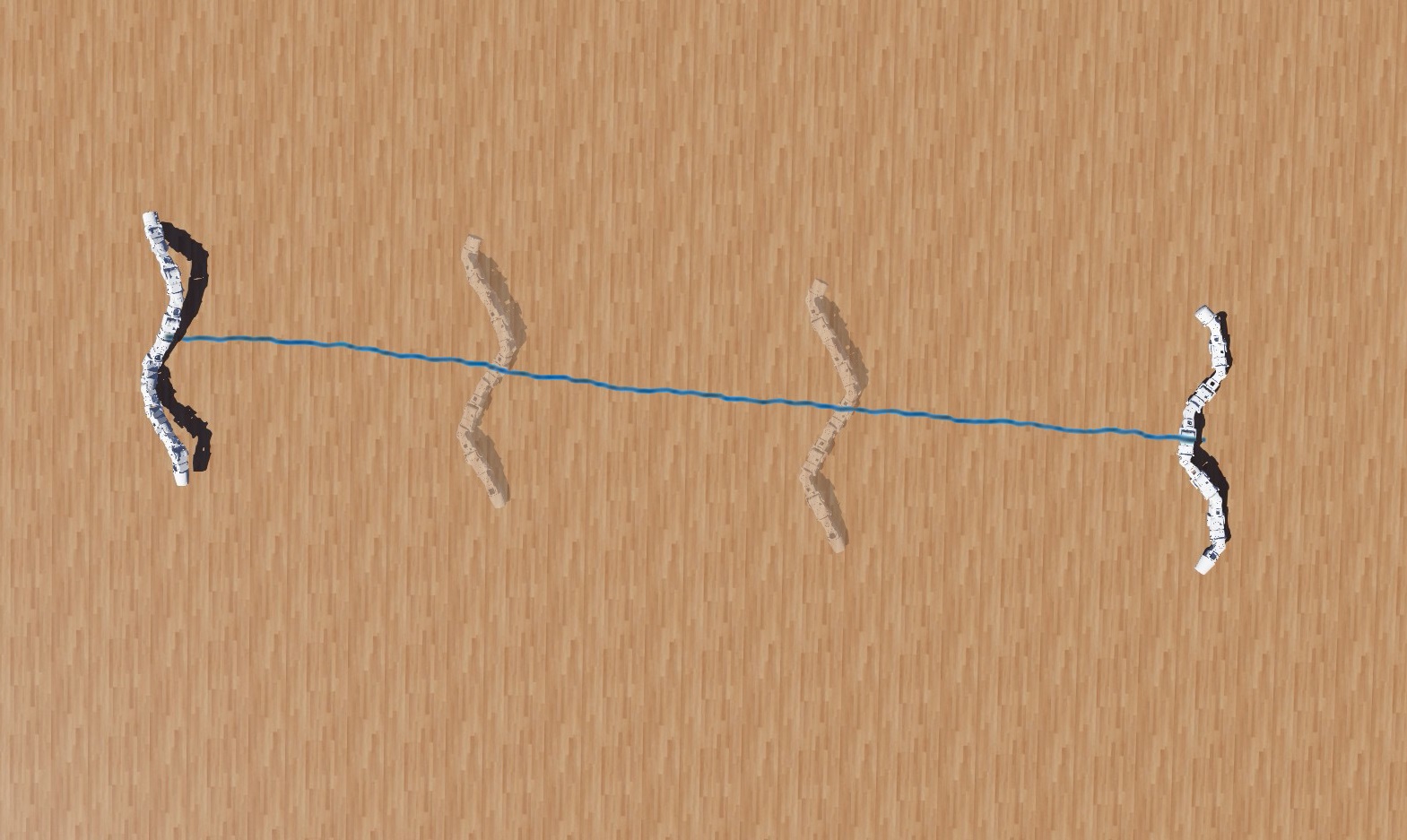}
    \caption{CoM trajectory of sidewinding.}
    \label{fig_21}
\end{figure}

\begin{figure}[t]
    \centering
    \includegraphics[width=1\linewidth]{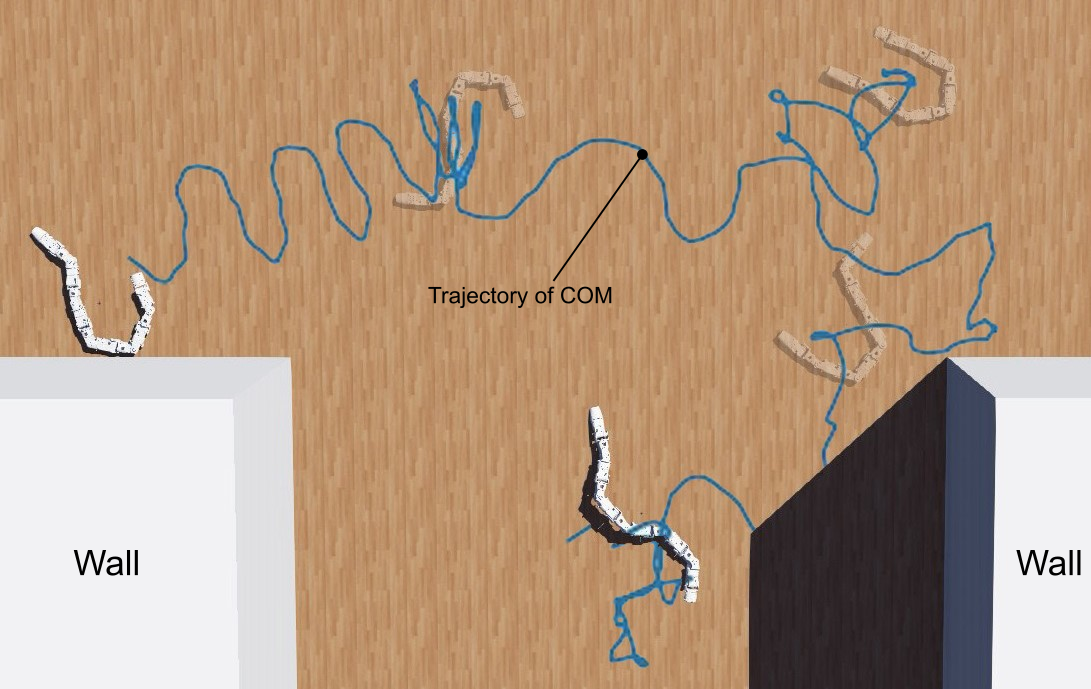}
    \caption{CoM trajectory of corner turning.}
    \label{fig_18}
\end{figure}

\begin{figure}[t]
\begin{subfigure}{.5\linewidth}
    \centering
    \includegraphics[height=.15\textheight]{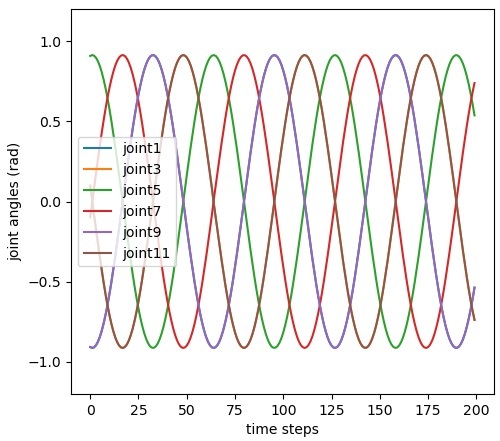}
    \caption{}
    \label{fig_20}
\end{subfigure}
\begin{subfigure}{.48\linewidth}
    \centering
    \includegraphics[height=.15\textheight]{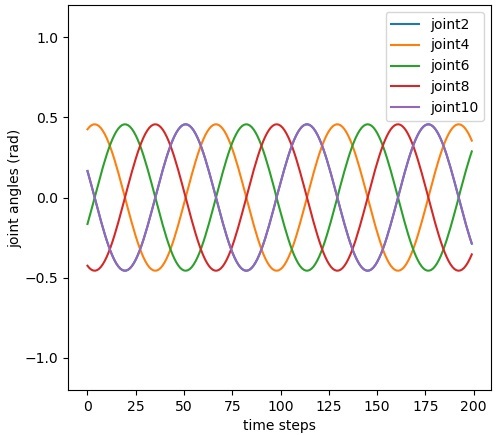}
    \caption{}
    \label{fig_19}
\end{subfigure}
\caption{Sidewinding: (a) pitch-axis joint trajectory; (b) yaw-axis joint trajectory.}
\end{figure}

\begin{figure}[t]
\begin{subfigure}{.5\linewidth}
    \centering
    \includegraphics[height=.15\textheight]{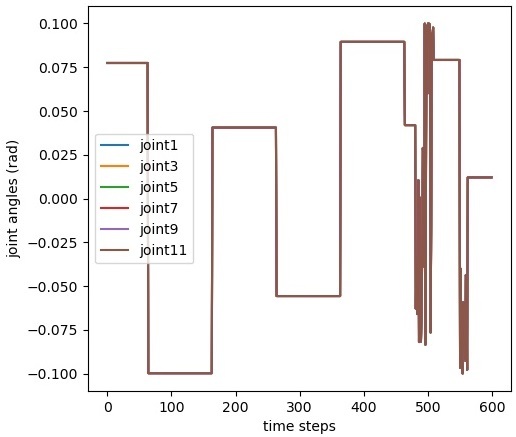}
    \caption{}
    \label{fig_12}
\end{subfigure}
\begin{subfigure}{.48\linewidth}
    \centering
    \includegraphics[height=.15\textheight]{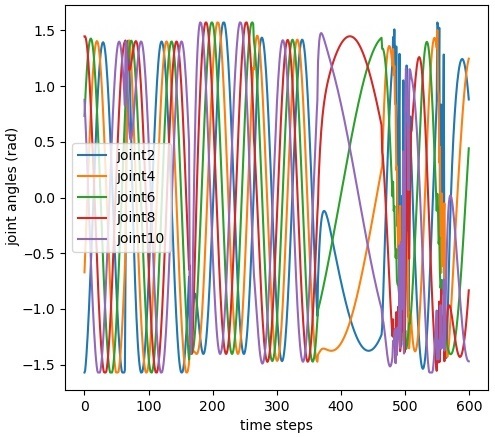}
    \caption{}
    \label{fig_13}
\end{subfigure}
\caption{Corner turning: (a) pitch-axis joint trajectory; (b) yaw-axis joint trajectory.}
\end{figure}

Our control method can be directly deployed in a new domain in a zero-shot manner, unlike traditional RL methods that need to be retrained for different environments. To demonstrate this capability, we chose a randomized maze map generated using Kruskal's algorithm (Fig.~\ref{fig_23}). COBRA spawns at a random location on the map and autonomously navigates to a randomly generated destination. Two crucial modes of motion, straight motion (by sidewinding) and turning (by corner turning), are shown in Figs.~ \ref{fig_21} and \ref{fig_18}. We also plot the trajectory of the CoM in Fig.~\ref{fig_21} and \ref{fig_18}. It can be seen that the turning motion is composed of multiple primitive motions, thus showing a more complex trajectory. The corresponding joint space movements are shown in Figs.~\ref{fig_20} to \ref{fig_13}. We plot the pitch-axis and yaw-axis movements independently. Readers can see that the two sets of joints that are perpendicular and independent of each other adopt different motion patterns to enable the robot to move. Still, the motion patterns are all composed of sinusoidal curves.

\begin{figure}[t]
    \centering
    \includegraphics[width=.9\linewidth]{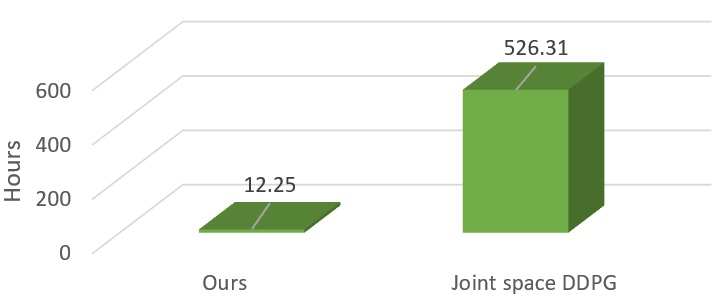}
    \caption{Training time comparison.}
    \label{fig_22}
\end{figure}

To show that the RL-CPG approach is an efficient scheme compared to applying RL directly in joint space (without CPG component), we compare the training time of both schemes on the same workstation (Fig.~\ref{fig_22}) with 3M running steps. We can see that the RL-CPG method significantly reduces the training time.


\section{Conclusion}
Classical joint space control design of slithering robots faces scalability challenges in large-scale navigation scenarios. In this paper, we design a four-layer motion control scheme for these robots. We tested the approach on the simulator of Northeastern's slithering robot, COBRA, to demonstrate its effectiveness. COBRA's model utilized the RL-CPG method for gait control and tree search for global navigation in complex maze-shaped environments. This design method not only dramatically shortens the training time of RL but also solves the problem of transferring the RL method to new environments.

\nocite{salagame_how_2023}
\printbibliography

@article{sihite_multi-modal_2023,
	title = {Multi-{Modal} {Mobility} {Morphobot} ({M4}) with appendage repurposing for locomotion plasticity enhancement},
	volume = {14},
	copyright = {2023 The Author(s)},
	issn = {2041-1723},
	url = {https://www.nature.com/articles/s41467-023-39018-y},
	doi = {10.1038/s41467-023-39018-y},
	language = {en},
	number = {1},
	urldate = {2023-07-04},
	journal = {Nature Communications},
	author = {Sihite, Eric and Kalantari, Arash and Nemovi, Reza and Ramezani, Alireza and Gharib, Morteza},
	month = jun,
	year = {2023},
	note = {Number: 1
Publisher: Nature Publishing Group},
	pages = {3323},
}

@inproceedings{hannigan_automatic_2020,
	title = {Automatic {Snake} {Gait} {Generation} {Using} {Model} {Predictive} {Control}},
	doi = {10.1109/ICRA40945.2020.9196853},
	booktitle = {2020 {IEEE} {International} {Conference} on {Robotics} and {Automation} ({ICRA})},
	author = {Hannigan, Emily and Song, Bing and Khandate, Gagan and Haas-Heger, Maximilian and Yin, Ji and Ciocarlie, Matei},
	month = may,
	year = {2020},
	note = {ISSN: 2577-087X},
	pages = {5101--5107},
}

@article{burdick_sidewinding_1994,
	title = {A 'sidewinding' locomotion gait for hyper-redundant robots},
	volume = {9},
	issn = {0169-1864},
	url = {https://doi.org/10.1163/156855395X00166},
	doi = {10.1163/156855395X00166},
	number = {3},
	urldate = {2023-01-11},
	journal = {Advanced Robotics},
	author = {Burdick, J.W. and Radford, J. and Chirikjian, G.S.},
	month = jan,
	year = {1994},
	note = {Publisher: Taylor \& Francis
\_eprint: https://doi.org/10.1163/156855395X00166},
	pages = {195--216},
}

@article{fukushima_model_2021,
	title = {Model {Predictive} {Path}-{Following} {Control} of {Snake} {Robots} {Using} an {Averaged} {Model}},
	volume = {29},
	issn = {1558-0865},
	doi = {10.1109/TCST.2020.3043446},
	number = {6},
	journal = {IEEE Transactions on Control Systems Technology},
	author = {Fukushima, Hiroaki and Yanagiya, Taro and Ota, Yusuke and Katsumoto, Masahiro and Matsuno, Fumitoshi},
	month = nov,
	year = {2021},
	note = {Conference Name: IEEE Transactions on Control Systems Technology},
	pages = {2444--2456},
}

@article{liljeback_modular_2005,
	series = {16th {IFAC} {World} {Congress}},
	title = {{MODULAR} {PNEUMATIC} {SNAKE} {ROBOT} {3D} {MODELLING}, {IMPLEMENTATION} {AND} {CONTROL}},
	volume = {38},
	issn = {1474-6670},
	url = {https://www.sciencedirect.com/science/article/pii/S147466701637286X},
	doi = {10.3182/20050703-6-CZ-1902.01274},
	language = {en},
	number = {1},
	urldate = {2023-01-11},
	journal = {IFAC Proceedings Volumes},
	author = {Liljebäck, Pål and Stavdahl, Øyvind and Pettersen, Kristin Y.},
	month = jan,
	year = {2005},
	pages = {19--24},
}

@inproceedings{ma_analysis_1999,
	title = {Analysis of snake movement forms for realization of snake-like robots},
	volume = {4},
	doi = {10.1109/ROBOT.1999.774054},
	booktitle = {Proceedings 1999 {IEEE} {International} {Conference} on {Robotics} and {Automation} ({Cat}. {No}.{99CH36288C})},
	author = {Ma, S.},
	month = may,
	year = {1999},
	note = {ISSN: 1050-4729},
	pages = {3007--3013 vol.4},
}

@inproceedings{ma_analysis_2003,
	title = {Analysis of creeping locomotion of a snake robot on a slope},
	volume = {2},
	doi = {10.1109/ROBOT.2003.1241899},
	booktitle = {2003 {IEEE} {International} {Conference} on {Robotics} and {Automation} ({Cat}. {No}.{03CH37422})},
	author = {Ma, Shugen and Tadokoro, N. and Li, Bin and Inoue, K.},
	month = sep,
	year = {2003},
	note = {ISSN: 1050-4729},
	pages = {2073--2078 vol.2},
}

@inproceedings{marafioti_study_2014,
	title = {A study of {Nonlinear} {Model} {Predictive} {Control} ({NMPC}) for snake robot path following},
	doi = {10.1109/ROBIO.2014.7090391},
	booktitle = {2014 {IEEE} {International} {Conference} on {Robotics} and {Biomimetics} ({ROBIO} 2014)},
	author = {Marafioti, Giancarlo and Liljebäck, Pål and Transeth, Aksel Andreas},
	month = dec,
	year = {2014},
	pages = {568--573},
}

@article{muller_economic_2020,
	series = {21st {IFAC} {World} {Congress}},
	title = {Economic model predictive control for obstacle-aided snake robot locomotion},
	volume = {53},
	issn = {2405-8963},
	url = {https://www.sciencedirect.com/science/article/pii/S2405896320333814},
	doi = {10.1016/j.ifacol.2020.12.2622},
	language = {en},
	number = {2},
	urldate = {2023-02-13},
	journal = {IFAC-PapersOnLine},
	author = {Müller, Evan and Köhler, Philipp N. and Pettersen, Kristin Y. and Allgöwer, Frank},
	month = jan,
	year = {2020},
	pages = {9702--9708},
}

@inproceedings{nonhoff_economic_2019,
	title = {Economic model predictive control for snake robot locomotion},
	doi = {10.1109/CDC40024.2019.9029627},
	booktitle = {2019 {IEEE} 58th {Conference} on {Decision} and {Control} ({CDC})},
	author = {Nonhoff, Marko and Köhler, Philipp N. and Kohl, Anna M. and Pettersen, Kristin Y. and Allgöwer, Frank},
	month = dec,
	year = {2019},
	note = {ISSN: 2576-2370},
	pages = {8329--8334},
}

@inproceedings{ohno_design_2001,
	title = {Design of slim slime robot and its gait of locomotion},
	volume = {2},
	doi = {10.1109/IROS.2001.976252},
	booktitle = {Proceedings 2001 {IEEE}/{RSJ} {International} {Conference} on {Intelligent} {Robots} and {Systems}. {Expanding} the {Societal} {Role} of {Robotics} in the the {Next} {Millennium} ({Cat}. {No}.{01CH37180})},
	author = {Ohno, H. and Hirose, S.},
	month = oct,
	year = {2001},
	pages = {707--715 vol.2},
}

@article{rincon_ver-vite_2003,
	title = {Ver-vite: dynamic and experimental analysis for inchwormlike biomimetic robots},
	volume = {10},
	issn = {1558-223X},
	shorttitle = {Ver-vite},
	doi = {10.1109/MRA.2003.1256298},
	number = {4},
	journal = {IEEE Robotics \& Automation Magazine},
	author = {Rincon, D.M. and Sotelo, J.},
	month = dec,
	year = {2003},
	note = {Conference Name: IEEE Robotics \& Automation Magazine},
	pages = {53--57},
}

@article{shan_design_1993,
	title = {Design and motion planning of a mechanical snake},
	volume = {23},
	issn = {2168-2909},
	doi = {10.1109/21.247890},
	number = {4},
	journal = {IEEE Transactions on Systems, Man, and Cybernetics},
	author = {Shan, Y. and Koren, Y.},
	month = jul,
	year = {1993},
	note = {Conference Name: IEEE Transactions on Systems, Man, and Cybernetics},
	pages = {1091--1100},
}

@incollection{umetani_biomechanical_1974,
	address = {Vienna},
	series = {International {Centre} for {Mechanical} {Sciences}},
	title = {Biomechanical {Study} of {Serpentine} {Locomotion}},
	isbn = {978-3-7091-2993-7},
	url = {https://doi.org/10.1007/978-3-7091-2993-7_12},
	language = {en},
	urldate = {2023-01-10},
	booktitle = {On {Theory} and {Practice} of {Robots} and {Manipulators}: {Volume} {I}},
	publisher = {Springer},
	author = {Umetani, Yoji and Hirose, Sigeo},
	editor = {Serafini, Paolo and Guazzelli, Elisabeth and Schrefler, Bernhard and Pfeiffer, Friedrich and Rammerstorfer, Franz G.},
	year = {1974},
	doi = {10.1007/978-3-7091-2993-7_12},
	pages = {171--184},
}

@inproceedings{wiriyacharoensunthorn_analysis_2002,
	title = {Analysis and design of a multi-link mobile robot ({Serpentine})},
	volume = {2},
	doi = {10.1109/ICIT.2002.1189249},
	booktitle = {2002 {IEEE} {International} {Conference} on {Industrial} {Technology}, 2002. {IEEE} {ICIT} '02.},
	author = {Wiriyacharoensunthorn, P. and Laowattana, S.},
	month = dec,
	year = {2002},
	pages = {694--699 vol.2},
}

@inproceedings{yim_new_1994,
	title = {New locomotion gaits},
	doi = {10.1109/ROBOT.1994.351134},
	booktitle = {Proceedings of the 1994 {IEEE} {International} {Conference} on {Robotics} and {Automation}},
	author = {Yim, M.},
	month = may,
	year = {1994},
	pages = {2508--2514 vol.3},
}

@inproceedings{ramezani_towards_2020,
	title = {Towards biomimicry of a bat-style perching maneuver on structures: the manipulation of inertial dynamics},
	shorttitle = {Towards biomimicry of a bat-style perching maneuver on structures},
	doi = {10.1109/ICRA40945.2020.9197376},
	booktitle = {2020 {IEEE} {International} {Conference} on {Robotics} and {Automation} ({ICRA})},
	author = {Ramezani, Alireza},
	month = may,
	year = {2020},
	note = {ISSN: 2577-087X},
	pages = {7015--7021},
}

@inproceedings{de_oliveira_thruster-assisted_2020,
	title = {Thruster-assisted {Center} {Manifold} {Shaping} in {Bipedal} {Legged} {Locomotion}},
	doi = {10.1109/AIM43001.2020.9158967},
	booktitle = {2020 {IEEE}/{ASME} {International} {Conference} on {Advanced} {Intelligent} {Mechatronics} ({AIM})},
	author = {de Oliveira, Arthur C. B. and Ramezani, Alireza},
	month = jul,
	year = {2020},
	note = {ISSN: 2159-6255},
	pages = {508--513},
}

@inproceedings{ramezani_generative_2021,
	title = {Generative {Design} of {NU}’s {Husky} {Carbon}, {A} {Morpho}-{Functional}, {Legged} {Robot}},
	doi = {10.1109/ICRA48506.2021.9561196},
	booktitle = {2021 {IEEE} {International} {Conference} on {Robotics} and {Automation} ({ICRA})},
	author = {Ramezani, Alireza and Dangol, Pravin and Sihite, Eric and Lessieur, Andrew and Kelly, Peter},
	month = may,
	year = {2021},
	note = {ISSN: 2577-087X},
	pages = {4040--4046},
}

@inproceedings{sihite_unilateral_2021,
	title = {Unilateral {Ground} {Contact} {Force} {Regulations} in {Thruster}-{Assisted} {Legged} {Locomotion}},
	doi = {10.1109/AIM46487.2021.9517648},
	booktitle = {2021 {IEEE}/{ASME} {International} {Conference} on {Advanced} {Intelligent} {Mechatronics} ({AIM})},
	author = {Sihite, Eric and Dangol, Pravin and Ramezani, Alireza},
	month = jul,
	year = {2021},
	note = {ISSN: 2159-6255},
	pages = {389--395},
}

@inproceedings{sihite_unsteady_2022,
	title = {Unsteady aerodynamic modeling of {Aerobat} using lifting line theory and {Wagner}'s function},
	doi = {10.1109/IROS47612.2022.9982125},
	booktitle = {2022 {IEEE}/{RSJ} {International} {Conference} on {Intelligent} {Robots} and {Systems} ({IROS})},
	author = {Sihite, Eric and Ghanem, Paul and Salagame, Adarsh and Ramezani, Alireza},
	month = oct,
	year = {2022},
	note = {ISSN: 2153-0866},
	pages = {10493--10500},
}

@article{grizzle_progress_nodate,
	title = {Progress on {Controlling} {MARLO}, an {ATRIAS}-series {3D} {Underactuated} {Bipedal} {Robot}},
	language = {en},
	author = {Grizzle, J W and Ramezani, A and Buss, B and Griﬃn, B and Hamed, K Akbari and Galloway, K S},
}

@inproceedings{sihite_orientation_2021,
	title = {Orientation stabilization in a bioinspired bat-robot using integrated mechanical intelligence and control},
	volume = {11758},
	url = {https://www.spiedigitallibrary.org/conference-proceedings-of-spie/11758/1175805/Orientation-stabilization-in-a-bioinspired-bat-robot-using-integrated-mechanical/10.1117/12.2587894.full},
	doi = {10.1117/12.2587894},
	urldate = {2023-05-17},
	booktitle = {Unmanned {Systems} {Technology} {XXIII}},
	publisher = {SPIE},
	author = {Sihite, Eric and Lessieur, Andrew and Dangol, Pravin and Singhal, Akshath and Ramezani, Alireza},
	month = apr,
	year = {2021},
	pages = {12--20},
}

@inproceedings{lessieur_mechanical_2021,
	title = {Mechanical design and fabrication of a kinetic sculpture with application to bioinspired drone design},
	volume = {11758},
	url = {https://www.spiedigitallibrary.org/conference-proceedings-of-spie/11758/1175806/Mechanical-design-and-fabrication-of-a-kinetic-sculpture-with-application/10.1117/12.2587898.full},
	doi = {10.1117/12.2587898},
	urldate = {2023-05-17},
	booktitle = {Unmanned {Systems} {Technology} {XXIII}},
	publisher = {SPIE},
	author = {Lessieur, Andrew and Sihite, Eric and Dangol, Pravin and Singhal, Akshath and Ramezani, Alireza},
	month = apr,
	year = {2021},
	pages = {21--27},
}

@inproceedings{kent2022improved,
  title={Improved Performance of CPG Parameter Inference for Path-following Control of Legged Robots},
  author={Kent, Nathan D and Neiman, David and Travers, Matthew and Howard, Thomas M},
  booktitle={2022 IEEE/RSJ International Conference on Intelligent Robots and Systems (IROS)},
  pages={11963--11970},
  year={2022},
  organization={IEEE}
}

@article{wugiasoihf,
  title={https://www.nasa.gov/feature/northeastern-university-slithers-to-the-top-with-big-idea-alternative-rover-concept}
}

@article{liu2021learning,
  title={Learning Contact-aware CPG-based Locomotion in a Soft Snake Robot},
  author={Liu, Xuan and Onal, Cagdas and Fu, Jie},
  journal={arXiv preprint arXiv:2105.04608},
  year={2021}
}

@article{liu2022reinforcement,
  title={Reinforcement Learning of a CPG-regulated Locomotion Controller for a Soft Snake Robot},
  author={Liu, Xuan and Onal, Cagdas and Fu, Jie},
  journal={arXiv preprint arXiv:2207.04899},
  year={2022}
}

@article{bellegarda2022cpg,
  title={CPG-RL: Learning central pattern generators for quadruped locomotion},
  author={Bellegarda, Guillaume and Ijspeert, Auke},
  journal={IEEE Robotics and Automation Letters},
  volume={7},
  number={4},
  pages={12547--12554},
  year={2022},
  publisher={IEEE}
}

@article{bing2020energy,
  title={Energy-efficient and damage-recovery slithering gait design for a snake-like robot based on reinforcement learning and inverse reinforcement learning},
  author={Bing, Zhenshan and Lemke, Christian and Cheng, Long and Huang, Kai and Knoll, Alois},
  journal={Neural Networks},
  volume={129},
  pages={323--333},
  year={2020},
  publisher={Elsevier}
}

@article{jia2021coach,
  title={A coach-based bayesian reinforcement learning method for snake robot control},
  author={Jia, Yuanyuan and Ma, Shugen},
  journal={IEEE Robotics and Automation Letters},
  volume={6},
  number={2},
  pages={2319--2326},
  year={2021},
  publisher={IEEE}
}

@article{shi2020deep,
  title={Deep reinforcement learning for snake robot locomotion},
  author={Shi, Junyao and Dear, Tony and Kelly, Scott David},
  journal={IFAC-PapersOnLine},
  volume={53},
  number={2},
  pages={9688--9695},
  year={2020},
  publisher={Elsevier}
}

@article{bellegarda2022visual,
  title={Visual CPG-RL: Learning Central Pattern Generators for Visually-Guided Quadruped Navigation},
  author={Bellegarda, Guillaume and Ijspeert, Auke},
  journal={arXiv preprint arXiv:2212.14400},
  year={2022}
}

@article{hasanzadeh2010ground,
  title={Ground adaptive and optimized locomotion of snake robot moving with a novel gait},
  author={Hasanzadeh, Shahir and Tootoonchi, Ali Akbarzadeh},
  journal={Autonomous Robots},
  volume={28},
  pages={457--470},
  year={2010},
  publisher={Springer}
}

@article{liljeback2011experimental,
  title={Experimental investigation of obstacle-aided locomotion with a snake robot},
  author={Liljeback, P{\aa}l and Pettersen, Kristin Y and Stavdahl, {\O}yvind and Gravdahl, Jan Tommy},
  journal={IEEE Transactions on Robotics},
  volume={27},
  number={4},
  pages={792--800},
  year={2011},
  publisher={IEEE}
}

@article{transeth20083,
  title={3-D snake robot motion: nonsmooth modeling, simulations, and experiments},
  author={Transeth, Aksel A and Leine, Remco I and Glocker, Christoph and Pettersen, Kristin Y},
  journal={IEEE transactions on robotics},
  volume={24},
  number={2},
  pages={361--376},
  year={2008},
  publisher={IEEE}
}

@inproceedings{bing2017cpg,
  title={CPG-based control of smooth transition for body shape and locomotion speed of a snake-like robot},
  author={Bing, Zhenshan and Cheng, Long and Huang, Kai and Zhou, Mingchuan and Knoll, Alois},
  booktitle={2017 IEEE International Conference on Robotics and Automation (ICRA)},
  pages={4146--4153},
  year={2017},
  organization={IEEE}
}

@inproceedings{liu2020learning,
  title={Learning to locomote with artificial neural-network and cpg-based control in a soft snake robot},
  author={Liu, Xuan and Gasoto, Renato and Jiang, Ziyi and Onal, Cagdas and Fu, Jie},
  booktitle={2020 IEEE/RSJ International Conference on Intelligent Robots and Systems (IROS)},
  pages={7758--7765},
  year={2020},
  organization={IEEE}
}

@article{lillicrap2015continuous,
  title={Continuous control with deep reinforcement learning},
  author={Lillicrap, Timothy P and Hunt, Jonathan J and Pritzel, Alexander and Heess, Nicolas and Erez, Tom and Tassa, Yuval and Silver, David and Wierstra, Daan},
  journal={arXiv preprint arXiv:1509.02971},
  year={2015}
}

@misc{salagame_how_2023,
	title = {How {Strong} a {Kick} {Should} be to {Topple} {Northeastern}'s {Tumbling} {Robot}?},
	url = {http://arxiv.org/abs/2311.14878},
	doi = {10.48550/arXiv.2311.14878},
	urldate = {2023-12-02},
	publisher = {arXiv},
	author = {Salagame, Adarsh and Bhattachan, Neha and Caetano, Andre and McCarthy, Ian and Noyes, Henry and Petersen, Brandon and Qiu, Alexander and Schroeter, Matthew and Smithwick, Nolan and Sroka, Konrad and Widjaja, Jason and Bohra, Yash and Venkatesh, Kaushik and Gangaraju, Kruthika and Ghanem, Paul and Mandralis, Ioannis and Sihite, Eric and Kalantari, Arash and Ramezani, Alireza},
	month = nov,
	year = {2023},
	note = {arXiv:2311.14878 [cs, eess]},
}

@article{bing_towards_2017,
	title = {Towards autonomous locomotion: {CPG}-based control of smooth {3D} slithering gait transition of a snake-like robot},
	volume = {12},
	issn = {1748-3190},
	shorttitle = {Towards autonomous locomotion},
	url = {https://dx.doi.org/10.1088/1748-3190/aa644c},
	doi = {10.1088/1748-3190/aa644c},
	language = {en},
	number = {3},
	urldate = {2023-07-07},
	journal = {Bioinspiration \& Biomimetics},
	author = {Bing, Zhenshan and Cheng, Long and Chen, Guang and Röhrbein, Florian and Huang, Kai and Knoll, Alois},
	month = apr,
	year = {2017},
	note = {Publisher: IOP Publishing},
	pages = {035001},
}

@inproceedings{nor_cpg-based_2014,
	title = {{CPG}-based locomotion control of a snake-like robot for obstacle avoidance},
	doi = {10.1109/ICRA.2014.6906634},
	booktitle = {2014 {IEEE} {International} {Conference} on {Robotics} and {Automation} ({ICRA})},
	author = {Nor, Norzalilah Mohamad and Ma, Shugen},
	month = may,
	year = {2014},
	note = {ISSN: 1050-4729},
	pages = {347--352},
}

@inproceedings{yang_hierarchical_2012,
	title = {A hierarchical connectionist {CPG} controller for controlling the snake-like robot's 3-dimensional gaits},
	doi = {10.1109/IROS.2012.6385578},
	booktitle = {2012 {IEEE}/{RSJ} {International} {Conference} on {Intelligent} {Robots} and {Systems}},
	author = {Yang, Guizhi and Ma, Shugen and Li, Bin and Wang, Minghui},
	month = oct,
	year = {2012},
	note = {ISSN: 2153-0866},
	pages = {822--827},
}

@article{wang_cpg-inspired_2017,
	title = {{CPG}-{Inspired} {Locomotion} {Control} for a {Snake} {Robot} {Basing} on {Nonlinear} {Oscillators}},
	volume = {85},
	issn = {1573-0409},
	url = {https://doi.org/10.1007/s10846-016-0373-9},
	doi = {10.1007/s10846-016-0373-9},
	language = {en},
	number = {2},
	urldate = {2023-07-03},
	journal = {Journal of Intelligent \& Robotic Systems},
	author = {Wang, Zhelong and Gao, Qin and Zhao, Hongyu},
	month = feb,
	year = {2017},
	pages = {209--227},
}

@article{manzoor_neural_2019,
	title = {Neural {Oscillator} {Based} {CPG} for {Various} {Rhythmic} {Motions} of {Modular} {Snake} {Robot} with {Active} {Joints}},
	volume = {94},
	issn = {1573-0409},
	url = {https://doi.org/10.1007/s10846-018-0864-y},
	doi = {10.1007/s10846-018-0864-y},
	language = {en},
	number = {3},
	urldate = {2023-03-13},
	journal = {Journal of Intelligent \& Robotic Systems},
	author = {Manzoor, Sajjad and Cho, Young Gil and Choi, Youngjin},
	month = jun,
	year = {2019},
	pages = {641--654},
}

\end{document}